\documentclass[journal]{IEEEtran}
\usepackage[T1]{fontenc}
\usepackage[utf8]{inputenc}
\usepackage{graphicx}
\usepackage{cite}
\usepackage{url}

\begin{document}

\title{When Multi-Robot Systems Meet Agentic AI:\\
Towards Embodied Collective Intelligence}

\author{Yuxuan~Yan, Yuanyuan~Jia, and Qianqian~Yang%
\thanks{The authors are with the College of Information Science and Electronic Engineering, Zhejiang University, Hangzhou, China (e-mail: \texttt{yanyx44@zju.edu.cn}; \texttt{yuanyuanjia@zju.edu.cn}; \texttt{qianqianyang20@zju.edu.cn}).}%
\thanks{This work is partly supported by the National Key R\&D Program of China under Grant No. 2024YFE0200802, by the National Natural Science Foundation of China (NSFC) under Grant Nos. 62293481, 62571487, and 62201505, and by the Zhejiang Provincial Natural Science Foundation of China under Grant No. LZ25F010001. (Corresponding author: Qianqian Yang.)}}

\maketitle

\begin{abstract}
Embodied AI is increasingly becoming agentic, shifting robots from perception--control pipelines towards closed-loop systems that can retrieve context, deliberate during execution, monitor feedback, and refine future behavior. In parallel, robotics research has also moved from single-robot autonomy towards multi-robot systems, driven by the need for wider sensing, distributed action, heterogeneous capabilities, and fault tolerance. As AI agents move from single-agent use towards multi-agent collaboration, robotics faces a parallel challenge: robot teams must move beyond sharing maps, task assignments, and datasets towards sharing the state produced by embodied agent loops. This article explores Embodied Collective Intelligence (ECI), a future multi-robot paradigm in which a robot team accumulates and uses world context, task progress, and skill experience as shared resources. Specifically, we first review how embodied AI is becoming agentic and how multi-robot cooperation has evolved. We then present Embodied Collective Intelligence through Co-Perception, Co-Action, and Co-Evolution. Finally, we use an illustrative navigation study to examine one concrete component of the concept: shared world-memory inheritance. The study shows that a newly added robot can benefit from merged team memory, but it is not intended as a full evaluation of the ECI framework. Taken together, the review and conceptual framework motivate Embodied Collective Intelligence as a direction for embodied multi-agent intelligence, while the case study grounds one measurable part of the concept.
\end{abstract}


\section{Introduction}
\IEEEPARstart{E}{mbodied} artificial intelligence (AI) is gradually moving out of simulators and controlled laboratories into workspaces that are closer to real deployment. In many benchmarks, a robot faces a fixed and clearly specified task: it completes one navigation, manipulation, or interaction episode, the environment and robot state are reset, and the next episode begins. This setting is valuable because it makes individual capabilities measurable. It is also far from how robots will be used in the world. A deployed robot often works repeatedly in the same space, while that space does not wait for the robot to finish. People move objects, doors open and close, corridors become temporarily blocked, and the robot's own actions may change the conditions of later tasks. Over time, the robot is not only facing the scene in front of its sensors, but also a changing workplace and the work history it accumulates inside it.

Responding only to the current observation is not enough for this kind of setting. Earlier robot systems were often organized as a pipeline from perception to control: sense the scene, estimate the state, plan an action, and execute it through a controller. This organization remains effective when the task boundary is clear. It becomes less sufficient when each task begins in a world already shaped by previous activity. The robot then has to compare what it sees now with what it has seen before, decide which records are still trustworthy, and choose its next action accordingly. Large language models and vision--language models change the system organization at precisely this point. A robot can put instructions, observations, and memory into a shared reasoning process; it can test its assumptions while acting rather than simply follow a prearranged action sequence; and it can keep selected traces of success or failure for later use. Recent building-wide mobile manipulation systems already show this agent-loop pattern in early form~\cite{bumble}. The robot is no longer only a pipeline that executes preset actions. It becomes closer to a stateful embodied agent, carrying memories of where it has been, plans that are still unfolding, and lessons from previous failures. In most systems today, however, this state remains private, locked inside one robot's own memory and execution process.

\begin{figure*}[!ht]
\centering
\includegraphics[width=0.96\textwidth]{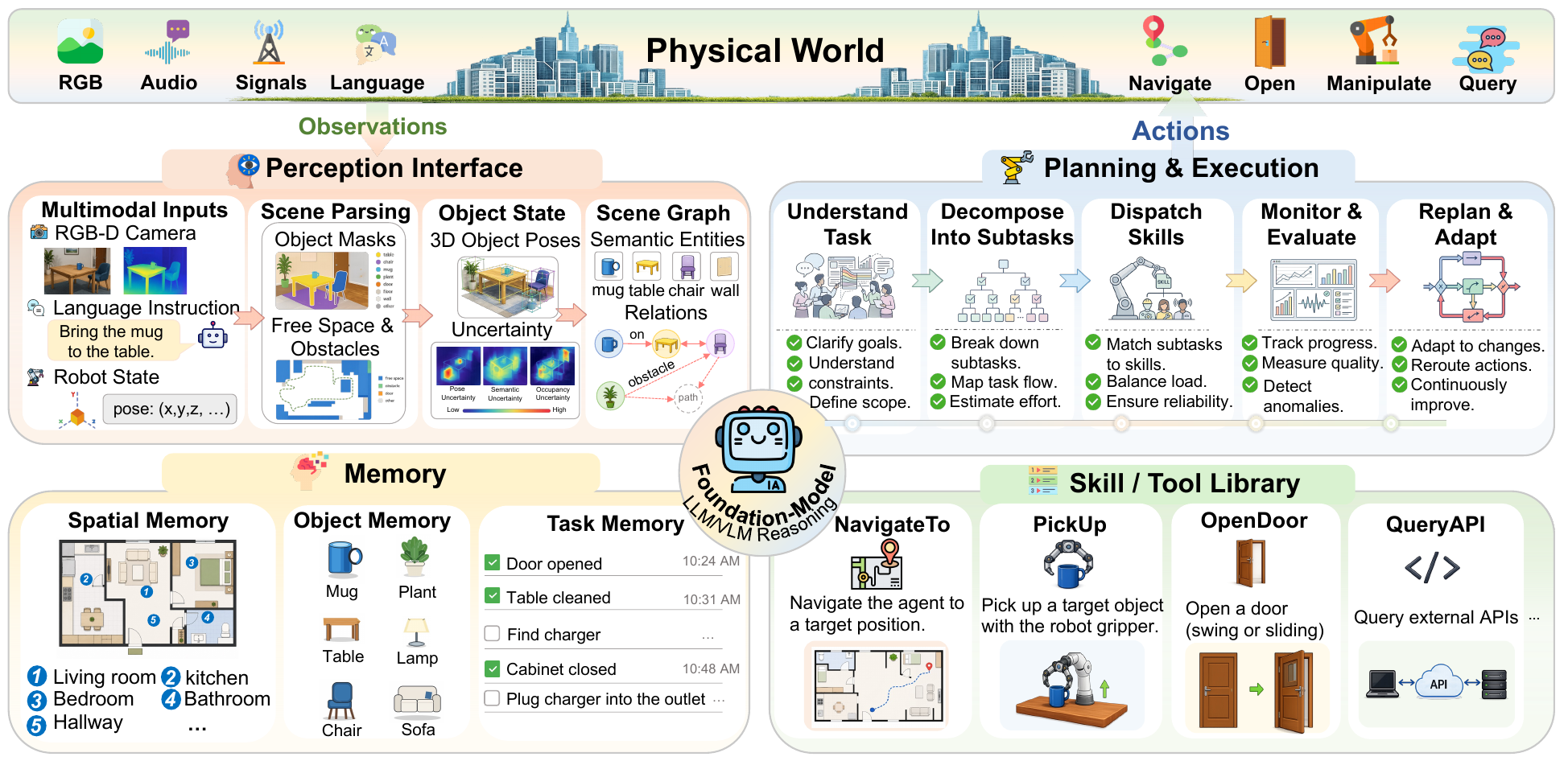}
\caption{A modular view of agentic embodied AI. Perception turns multimodal observations into structured context; memory stores spatial, object, and task information; the foundation-model brain reasons over these resources; planning/execution grounds decisions in action; and the skill/tool library provides callable embodied capabilities.}
\label{fig:agentframe}
\end{figure*}

At the same time, many real deployments are multi-robot by nature. Fleets of mobile platforms move through warehouses; delivery, cleaning, and inspection robots may share hospitals and office buildings; domestic spaces may gradually contain service robots with different bodies and functions. These robots may not be doing the same task, and they may not have the same capabilities. Yet they move through the same space and are affected by the same changes. If one robot has already explored a room, a newly arrived robot should not have to search it from scratch. If one robot has just found that a corridor is blocked, the next robot should not have to rediscover the same fact through another failed attempt.

Multi-robot research has built strong foundations for this kind of cooperation, especially in task allocation, path coordination, map fusion, shared knowledge bases, and collective learning. These mechanisms mainly exchange answers to questions such as who should do what and where things are. Once each robot starts to operate as an embodied agent, however, the team begins to need another kind of shared resource: the internal state that each robot updates while working. This includes when an observation was made, why a plan changed, under what conditions a failure occurred, and whether a useful experience can be continued by another robot. Existing sharing mechanisms do not yet fully carry this kind of agent-loop state.

This is where two lines of development meet. Individual robots are becoming more agentic, while real deployments increasingly involve many robots. The team therefore needs to share more than tasks and maps; it needs ways to share the state produced by embodied agents during operation. This article calls the resulting future multi-robot paradigm \emph{Embodied Collective Intelligence} (ECI). ECI does not connect all robots to a central brain, nor does it assume that the robots have identical bodies. Each robot still senses and acts through its own embodiment and remains responsible for local safety. What ECI adds is a shared layer for organizing observations, task progress, and skill experience as collective resources that can be read, validated, and inherited. Co-Perception turns distributed observations into team world memory. Co-Action records task progress and reassignment in a shared ledger. Co-Evolution consolidates local attempts into skill experience that can be reused according to embodiment. In short, ECI shares state rather than control: the team reads, writes, and cross-checks shared records, while concrete perception and action remain local.

This structure also makes plug-and-play robots more concrete. A new robot entering an environment where a team has already worked should not begin from an empty history. It can first read the team's world memory, understand which regions have been observed and which records were recently updated, inspect ongoing tasks, and retrieve experience compatible with its own capabilities before acting locally. Many problems remain open, including stale information, transfer across heterogeneous bodies, communication overhead, safety constraints, and responsibility boundaries. These problems are precisely why multi-robot research is shifting from scheduling machines towards building a collective capability among embodied agents.

This article makes three contributions:
\begin{enumerate}
\item \textbf{It reviews the agentification of embodied AI} (Section~\ref{sec:background}), focusing on how memory, planning, and self-evolution have begun to turn single robots from one-shot task executors into stateful embodied agents.
\item \textbf{It reexamines multi-robot cooperation through this agentic lens} (Section~\ref{sec:teamscale}), showing how task coordination, shared environmental representations, and shared learning provide foundations for embodied multi-agent collaboration.
\item \textbf{It proposes Embodied Collective Intelligence as a future multi-robot blueprint} (Sections~\ref{sec:framework}--\ref{sec:casestudy}), where Co-Perception, Co-Action, and Co-Evolution organize shared team state, and an illustrative target-object navigation study tests one measurable component: inherited world memory.
\end{enumerate}


\section{Agentic Embodied Intelligence}
\label{sec:background}
Agentic embodied AI does not simply mean placing a larger model on a robot. It means that embodied systems are borrowing the organization of AI agents: a capable foundation model becomes the reasoning core, while perception, memory, planning, execution, and self-evolution are reorganized around that core. This modular view is not the only form of foundation-model robotics. Vision--language--action policies may collapse perception, reasoning, and control into an end-to-end model. Here we focus on the agentic, hierarchical line because its components are explicit: an LLM or vision--language model (VLM) interprets the instruction, reasons about the scene, sketches a high-level plan, monitors progress, and sometimes reflects on failure~\cite{bumble,saycan}. The surrounding robotics modules then determine whether this reasoning can remain grounded in the physical world.

Fig.~\ref{fig:agentframe} sketches this common loop. Perception receives sensor streams such as RGB images, depth maps, proprioception, and language observations, then translates them into structures the brain can reason over. Memory retains part of this information as retrievable context about places, objects, and previous tasks. The foundation-model brain combines the task instruction, perceptual summaries, retrieved memory, and available embodied capabilities to reason about what should happen next.

Planning and execution turn this reasoning into grounded action: the task is decomposed, actions are dispatched to available robot behaviors, outcomes are monitored, and the plan is revised when feedback contradicts the agent's assumptions. Finally, selected execution traces may feed self-evolution, updating prompts, skill descriptions, code routines, policies, or verification rules for future behavior. The rest of this section reviews three agentic components in turn: memory as retrievable embodied context, planning as execution-time deliberation, and self-evolution as a constrained route from experience to improved behavior.

\subsection{Memory for embodied AI}
Memory is the retained context that an embodied agent can retrieve. In robotics, this idea is grounded less in language-style memory and more in mapping. It began as an occupancy grid, feature map, or metric reconstruction that supported localization, navigation, and coverage. As embodied tasks moved from reaching places to acting in realistic scenes, these maps became more semantic. Robots began to attach object labels, spatial relations, topology, timestamps, and task traces to the environment, producing semantic maps and open-vocabulary 3D scene graphs that a language-conditioned agent can query~\cite{semmapsurvey,conceptgraphs}.

The recent shift is therefore not that robots suddenly acquired memory, but that memory is becoming a retrieval interface for the foundation-model brain. A new instruction can be grounded against what was observed before, where objects were last seen, and which parts of the scene may need verification. The hard part is memory management. Real environments change: objects move, doors open or close, and old observations become misleading. Dynamic-memory systems such as DynaMem~\cite{dynamem} target this problem by making memory more time-aware and updateable, but they also reveal the current limit: embodied memory must constantly decide what to write, retrieve, refresh, or overwrite, and it is still mostly maintained inside a single robot.

\subsection{Planning for embodied AI}
Planning here refers to the high-level agentic process that turns a task goal into grounded decisions during execution. It is different from low-level motion planning or control, and it is also different from end-to-end action prediction in VLA policies. Earlier embodied systems often followed a plan-then-execute pattern: a task planner produced an action sequence from the initial state, and navigation or manipulation modules tried to carry it out. Feedback existed, but it was often local to control, recovery, or failure detection rather than part of a continuous high-level deliberation loop.

The foundation-model era changed the interface. Language models can interpret instructions, decompose goals, and reason over scene descriptions, while lower-level policies still handle motion, contact, and sensing. SayCan~\cite{saycan} made the grounding problem explicit by selecting language-plan steps according to what the robot could actually execute. Agentic planning pushes this further from one-shot sequencing towards execution-time deliberation: the robot monitors whether assumptions hold, uses memory or perception to re-check the scene, and revises the next step when execution deviates from the plan. The unresolved boundary is brittleness. The planner still depends on noisy perception, incomplete memory, and imperfect capability estimates, and its working state usually remains inside one robot's context window, planner memory, or execution log.

\subsection{Self-evolution for embodied AI}
Self-evolution is the update path from execution experience to future behavior. The term should be read carefully: in this article, it refers to constrained experience consolidation and reuse rather than open-ended autonomous evolution. In earlier embodied systems, skills were usually fixed policies, controllers, or primitives; a planner could select among them, but a success or failure rarely changed the robot's future repertoire. The most mature path for improvement has been data-driven and largely offline: robot experiences are collected, aggregated, and used to train broader policies, as illustrated by cross-embodiment efforts such as Open X-Embodiment~\cite{openx}.

Recent agentic systems add a lighter feedback path. Execution traces, visual feedback, and task outcomes can be summarized and used to revise prompts, code routines, skill descriptions, failure explanations, or future training data. REFLECT~\cite{reflect} represents an early form of this route by summarizing robot experiences for failure explanation and correction, while EmbodiSkill~\cite{embodiskill} updates embodied skills from execution evidence. These works show the direction, but also the limit. Current self-evolution mostly changes procedural descriptions, reflection records, or data used for later learning; acquiring genuinely new physical capabilities still requires data collection, retraining, and embodiment-specific verification.

\begin{figure*}[!ht]
\centering
\includegraphics[width=0.98\textwidth]{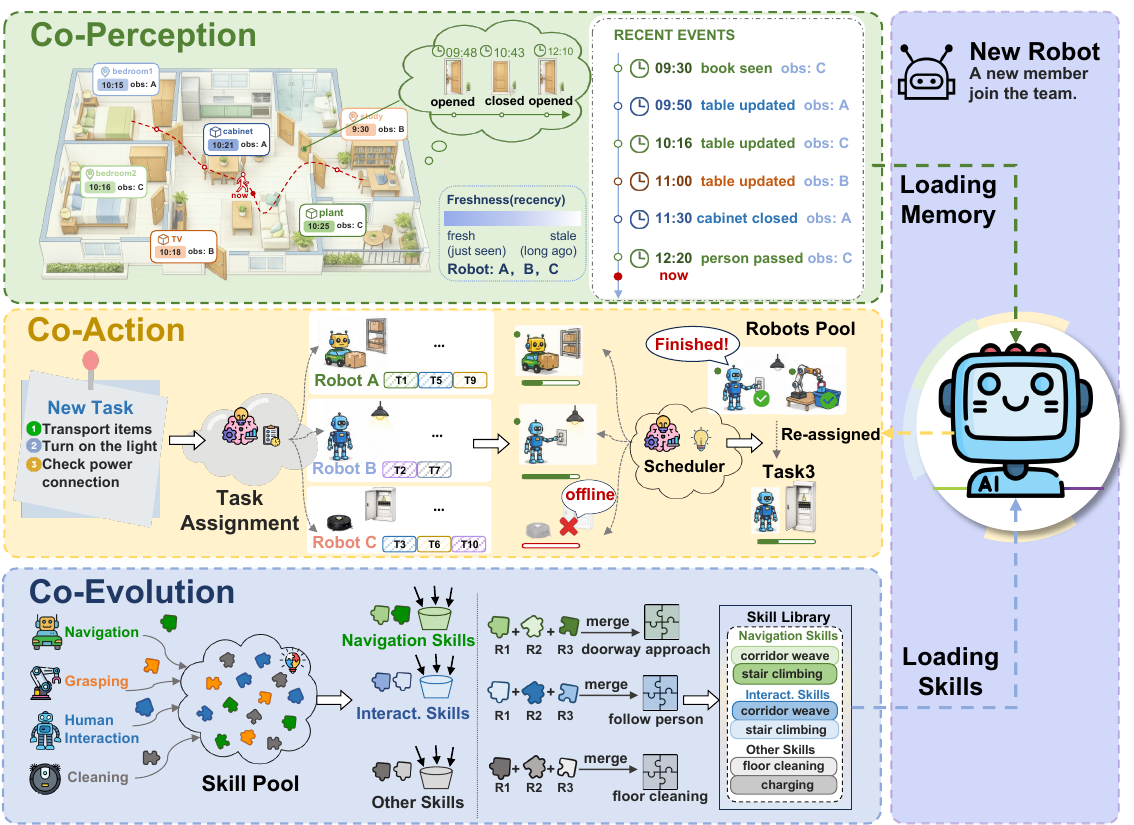}
\caption{Blueprint of Embodied Collective Intelligence. Co-Perception builds shared world memory from distributed observations. Co-Action maintains a task-state ledger for commitments, progress, failures, and reassignment. Co-Evolution consolidates local attempts into a skill library. A newly arrived robot reads these shared resources, loads compatible skills, and joins the ongoing team.}
\label{fig:concept}
\end{figure*}

\subsection{What agentification has delivered}
Step back and the pattern is simple. The agent paradigm is giving embodied AI a loop organized around a foundation-model brain and supported by memory, planning, and self-evolution. The robot can retrieve retained context, revise its plan while acting, and fold selected execution experience into future prompts, procedures, or policies. This is a major shift from a perception--control pipeline, but it should not be overstated. Most demonstrations remain system-specific, embodiment-bound, and largely single-robot. The next section asks what happens when several such agents must operate together.

\section{From Robot Cooperation to Multi-Agent Collaboration}
\label{sec:teamscale}
Multi-robot systems emerged from the physical limits of a single robot. One robot has finite sensing range, payload, battery life, speed, and fault tolerance; a team can cover larger spaces, divide tasks, observe from multiple viewpoints, and continue working when one member fails. The central problem has therefore been practical from the beginning: robots must coordinate under limited communication, partial observability, and heterogeneous capabilities. The answer has not been a clean succession of paradigms. It is better read as the accumulation of several sharing dimensions: robots share commitments about who will do what, representations of the environment, and experience for improving policies or skills. These dimensions echo the individual-agent capabilities discussed in Section~\ref{sec:background}, but in multi-robot research they have mostly developed as separate coordination mechanisms.

\subsection{Coordination of actions and commitments}
One foundational layer is action coordination. In swarm and distributed robot systems, individual robots often follow local rules, react to nearby peers, and produce useful group behavior through repeated interaction. This line established a basic lesson: coordination does not always require a central supervisor. Redundancy, local sensing, and simple interaction rules can produce robust behavior when tasks can be expressed through local interactions.

As tasks became more structured, cooperation also became an allocation problem. Robots had to decide who should explore which region, carry which object, or respond to which request. Market-based methods, including consensus-based bundle auctions~\cite{cbba}, made this layer explicit: robots bid, claim tasks, and converge on assignments without a single supervisor. What the team shared was mainly a ledger of commitments. It answered ``who does what now,'' but it did not by itself preserve rich context about what the team had observed or learned.

\subsection{Coordination through shared environmental representations}
Another layer made the environment itself the object of sharing. Collaborative SLAM and multi-robot mapping allowed several robots to fuse observations into a common geometric model. Systems such as Kimera-Multi~\cite{kimeramulti} pushed this line towards distributed metric--semantic mapping, where the shared representation began to include not only geometry but also object and scene-level meaning.

\begin{figure*}[!ht]
\centering
\includegraphics[width=0.95\textwidth]{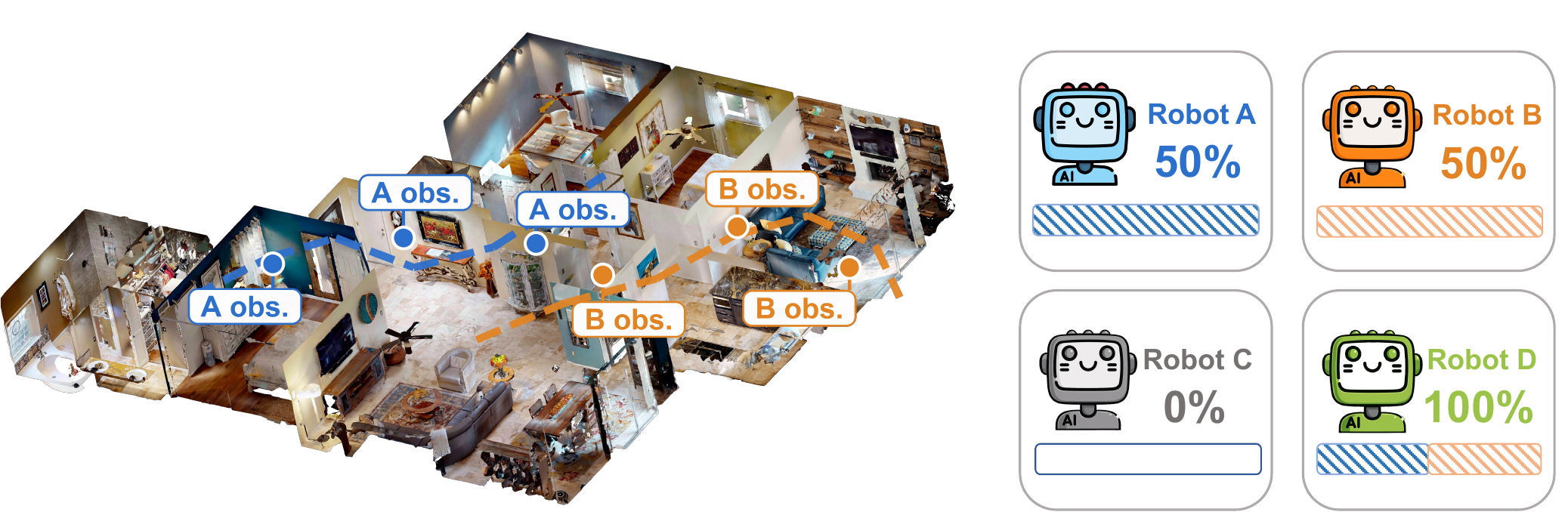}
\caption{Robot memory settings in the case study. Robots A and B each retain partial memories from their own prior trajectories, Robot C starts without prior memory, and Robot D inherits the merged memory from Robots A and B.}
\label{fig:robotsettings}
\end{figure*}

Cloud robotics extended this idea beyond one deployment. RoboEarth~\cite{roboearth} framed robot knowledge as something that could be uploaded, searched, and reused: maps, object models, and action recipes became shared resources rather than private artifacts. This introduced persistence and reuse, but the shared resource was still largely a repository. Robots could consult or update it, yet the representation did not necessarily capture the live intent, uncertainty, or recent failures of each robot.


\subsection{Coordination through shared learning}
Robot learning changed the unit of sharing again. Instead of sharing only assignments, maps, or symbolic recipes, robots could contribute demonstrations, trajectories, failures, and model updates. Cross-embodiment datasets such as Open X-Embodiment~\cite{openx} show the power of this layer: experience from many robots can train policies that generalize across platforms and tasks.

The limitation is that this sharing is usually offline and model-centric. A fleet may improve a policy, but the improvement often appears after data aggregation, training, and deployment. During a mission, the team may still lack a live account of what has just been learned, who can use it, and whether it changes the current plan. Learning-era sharing therefore made robot experience scalable, but not yet a real-time coordination mechanism.

\subsection{Towards embodied multi-agent collaboration}
Section~\ref{sec:background} described how a single embodied robot is becoming an agent: it can retrieve memory, plan during execution, and fold selected experience back into future behavior. This does not mean that today's multi-robot systems have already become collective agents. It suggests a narrower but important shift: as each robot becomes more agent-like, coordination may need to include the state of its agent loop, not only its assigned task, map contribution, or training data.

The opportunity is that the earlier sharing dimensions begin to meet. Environmental representations could become richer team context rather than static maps. Commitment ledgers could become task progress that reveals what each robot is trying, where it is blocked, and what needs reassignment. Shared learning could move closer to the mission by recording which recovery behaviors or local skills worked under which embodiment and environment conditions. These are still emerging possibilities rather than settled mechanisms, but they show why memory, planning, and self-evolution may matter beyond a single robot.

Recent systems provide early hints of this convergence. EMOS~\cite{emos} uses LLM agents to reason about heterogeneous robot embodiments and coordinate roles. RoboOS-NeXT~\cite{roboosnext} organizes heterogeneous robots around high-level reasoning, low-level skills, and shared memory. These works suggest that language and foundation models can help robots describe tasks, capabilities, and context in a common form. Yet they also show that the field is still early: coordination often depends on a platform-level coordinator, shared memory is not always written and read as a persistent team resource by every robot, and reusable experience remains hard to validate across different bodies. The next section turns this convergence into a design blueprint for organizing shared resources explicitly at the collective level.

\section{Embodied Collective Intelligence}
\label{sec:framework}
Sections~\ref{sec:background} and~\ref{sec:teamscale} point to the same missing layer. Individual robots are gaining agent loops that remember, plan, and improve, while robot teams have learned to share assignments, maps, and training data. Embodied Collective Intelligence is a blueprint for future multi-robot embodied intelligence that brings these two lines together. It does not ask robots to merge into one centralized super-agent. It asks how a robot team should organize the shareable records produced by many embodied agent loops: what the team has observed, what it is doing, and what experience can help future behavior.

Fig.~\ref{fig:concept} shows this multi-robot blueprint. The body-bound parts of intelligence stay local: sensing, motion, contact, safety checks, and low-level control. The shared layer stores compact records that can outlive a single execution: what has changed in the world, what work is currently underway, and what experience may help future behavior. Co-Perception turns distributed observations into world memory; Co-Action turns private execution into team-visible task progress; and Co-Evolution turns scattered attempts into a skill library. The shared layer matters because these records must be fresh enough for a running team, but shared beyond any one robot's onboard computer.

\subsection{Co-Perception: from many views to world memory}
Co-Perception corresponds to the top band of Fig.~\ref{fig:concept}. Robots A, B, and C observe different rooms, objects, and events. Each useful observation is written into the world memory: an object was seen, a door changed state, a table was moved, or a room was checked. A memory entry should therefore contain not only \emph{what} was observed, but also \emph{when} it was observed, \emph{who} observed it, and how fresh the observation remains. This is why the figure includes timestamps, observer labels, a freshness indicator, and a recent-events list.

The goal is not to build a perfect global model of the world. It is to give the team a shared, queryable account of the parts of the environment that matter for action. A robot looking for a mug should know that another robot saw it on the dining table ten minutes ago, and also that the same area has since been updated by a different robot. If a door was seen open, then closed, then open again, the event history tells the next robot whether to trust the old state or verify it. In communication terms, the team does not need to stream every camera frame; it needs to synchronize compact updates about changed objects, changed relations, and stale entries. Co-Perception therefore extends the memory line of Section~\ref{sec:background} from one robot's retained context to a team-level account of a changing environment.

\subsection{Co-Action: from private plans to team progress}
Co-Action corresponds to the middle band of Fig.~\ref{fig:concept}. A new task is decomposed into subtasks, assigned to available robots, and then monitored as execution unfolds. In a conventional assignment system, the team may know who was initially assigned to which task. In an agentic team, that is not enough. Plans change during execution: a robot may find a blocked corridor, fail to open a cabinet, finish early, or discover that another robot is better placed to continue. These changes need to be visible while the mission is still running.

The task-state ledger is the shared record of this present tense rather than an auction market. It contains open tasks, claims, progress, failures, released commitments, finished subtasks, and requests for help. A robot can still plan locally, but its commitments and updates become available to others. The scheduler shown in the figure is a logical coordination service rather than a central controller: it records task state, exposes conflicts, and triggers reassignment when a robot goes offline or when work has already been completed by another member. Co-Action therefore extends the planning line of Section~\ref{sec:background}: planning is no longer only a private chain of thoughts and actions, but a source of live coordination state for the team.

\subsection{Co-Evolution: from scattered attempts to shared skills}
Co-Evolution corresponds to the bottom band of Fig.~\ref{fig:concept}. Local attempts from navigation, grasping, human interaction, cleaning, and other tasks first enter a skill pool. A single entry is rarely a finished skill. It may be a partial recovery routine, a better parameter setting, a failure condition, or a useful composition of existing actions. By itself, such an entry can be noisy, narrow, or duplicated. Its value grows when the team can compare it with similar attempts from other robots.

The ``categorize and merge'' step in the figure is therefore essential. Skill fragments are grouped by task type, checked against their evidence, deduplicated when they describe the same behavior, and merged when several attempts point to a more reliable routine. Several robots may report different ways to approach a doorway, recover from a navigation failure, or grasp the same class of object. Co-Evolution asks which parts are redundant, which parts are embodiment-specific, and which parts should be kept as alternatives because different bodies or environments need different solutions.

The result is a skill library that a robot can load from, as shown on the right of the bottom band. Compatibility remains the boundary. A navigation recovery learned by a wheeled robot may help another wheeled robot immediately; a grasping routine may only help robots with a similar arm and gripper. The shared layer should therefore store not just the skill name, but the evidence behind it, the preconditions under which it worked, and the body conditions required to execute it. Each skill record should include task type, preconditions, embodiment requirements, execution interface, evidence count, success/failure contexts, and verification status. This is different from claiming that the team can train a universal policy online. Co-Evolution is a lighter and more immediate layer: it lets the team turn imperfect individual experience into reusable skill knowledge over time.

\subsection{Inheritance as the litmus test}
The simplest way to judge whether Embodied Collective Intelligence is genuinely collective is to ask what happens when a new robot arrives. In a conventional deployment, the newcomer may have capable policies and controllers, but it starts with little local history. It must discover the environment, learn which skills are useful there, and negotiate what is currently being done.

In Embodied Collective Intelligence, arrival begins with a read, as shown by the new robot on the right side of Fig.~\ref{fig:concept}. From the world memory, the newcomer learns what has recently been observed and what may be stale. From the task-state ledger, it sees open tasks, current commitments, and recent failures. From the skill library, it loads procedures that match its body. It still verifies locally before acting, and it cannot use a skill its body cannot execute. But it enters a team whose perception, action, and learning have already accumulated. That inheritance is the central promise of Embodied Collective Intelligence.

\begin{figure*}[!t]
\centering
\includegraphics[width=0.98\textwidth]{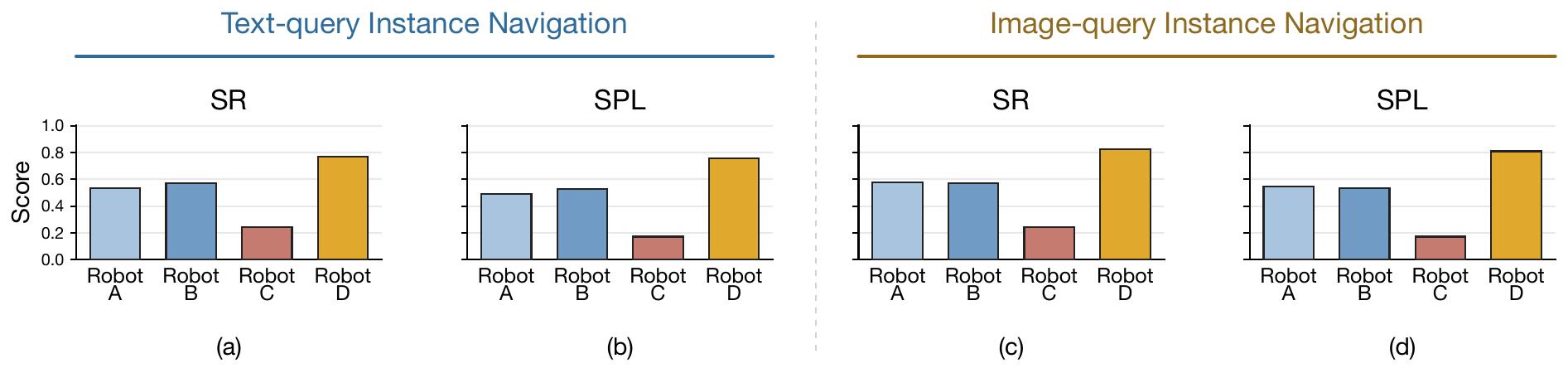}
\caption{World-memory inheritance in the target-object navigation case study. Panels (a) and (b) report SR and SPL for text-query instance navigation; panels (c) and (d) report SR and SPL for image-query instance navigation. Robots A and B use partial memories, Robot C has no memory, and Robot D reads the merged team memory.}
\label{fig:casestudy}
\end{figure*}

\section{Case Study}
\label{sec:casestudy}
The most direct way to make Embodied Collective Intelligence measurable is to test sharing and inheritance. We therefore use a deliberately narrow case study: only the world-memory part of the concept is instantiated, while Co-Action and Co-Evolution are kept outside the measurement. The test is simple: a new robot is asked to navigate to a target object, either with access to the team's explored world memory or by searching from scratch. Although narrow, target-object navigation is a representative substrate for embodied tasks: before a robot can grasp, inspect, deliver, or interact with an object, it often has to identify where that object is and reach it reliably.

We run the study on four indoor scenes from the Habitat--Matterport 3D dataset~\cite{hm3d}\footnote{\url{https://aihabitat.org/datasets/hm3d/}}, with 166 target-object navigation tasks in total. Each task names a target object that the robot must locate and reach. We define a robot's memory as an indexed record built from its past work trajectory. During previous exploration or task execution, the robot records RGB-D observations, poses, detected objects, and local semantic cues. Qwen3.5-Plus\footnote{\url{https://qwen.ai/blog?id=qwen3.5}}, used as the VLM-centered agent reasoner, analyzes these trajectories and builds an open-vocabulary semantic index over candidate object locations, visual evidence, and spatial nodes.

The four robots differ in their history rather than their body or policy. Robots~A and~B each have a continuous work trajectory in the same indoor space, with each trajectory covering roughly half of the environment. The two partial memories therefore come from comparable trajectory coverage rather than different robot capabilities. Robot~C is a newcomer with no historical memory and must begin from scratch. Robot~D is also a newcomer, but it inherits the shared memory obtained by merging the memories of Robots~A and~B, as illustrated in Fig.~\ref{fig:robotsettings}.


Each robot acts through the same agentic loop. Given a target, it first retrieves relevant entries from its own memory, including candidate locations and visual context. Qwen3.5-Plus then reasons over the current observation and the retrieved memory to choose between two actions: navigate towards a remembered location, or continue free exploration. In the latter case, the robot calls Uni-NaVid\footnote{\url{https://github.com/jzhzhang/NaVid-VLN-CE}} as the VLN model. After moving and receiving a new observation, the robot repeats this retrieve--reason--navigate/explore loop until it reaches the target or the trial ends.

The result is a clean inheritance effect. With no memory, Robot~C succeeds on only $24.1\%$ of tasks. Robots~A and~B, each holding half of the explored context, reach about $54$--$58\%$ success depending on the query mode. Robot~D, which inherits the merged team memory, reaches $77.1\%$ SR and $0.757$ SPL in text-query instance navigation, and $82.5\%$ SR and $0.809$ SPL in image-query instance navigation. The gain is not only from having more entries. Merging the two explored halves increases object and region coverage and, where trajectories overlap, can fuse repeated observations into more reliable spatial associations, giving the newcomer a more useful context than any single robot's private memory.

We report two metrics. Success rate (SR) counts a trial as successful when the robot stops within two meters of the target object. Success weighted by path length (SPL)~\cite{spl} further penalizes inefficient routes, so it measures not only whether the robot finds the target but also how efficiently it gets there. Under these metrics, Robot~D's SPL nearly tracks its success rate. By contrast, Robot~C's SPL is only $0.172$, well below its $0.241$ success rate, because from-scratch search often wanders even when it eventually succeeds. The same pattern appears in both text-query and image-query instance navigation. The study should be read as an illustrative measurement of memory inheritance, rather than a benchmark-level evaluation of embodied navigation. A team that shares memory lets a newcomer act from collective context; a robot without that inheritance behaves like an isolated explorer.

\section{Conclusion}
\label{sec:conclusion}
This article has framed Embodied Collective Intelligence as a future paradigm for multi-robot embodied intelligence. The motivation is simple: individual embodied robots are becoming agentic, but the memory, task progress, and experience produced by their agent loops still remain mostly private. Embodied Collective Intelligence lifts these records to the team level through three shared resources: world memory built by Co-Perception, task progress maintained by Co-Action, and skill knowledge grown by Co-Evolution. The illustrative target-object navigation study grounds one measurable part of this vision: by inheriting the team's merged world memory, a new robot can avoid much of the from-scratch search and act from collective context. The remaining challenge is to make such shared resources fresh, trustworthy, and reusable across heterogeneous bodies and constrained communication links. Progress on this shared layer would move robot teams away from isolated deployments and towards a more collective form of embodied intelligence.

\bibliographystyle{IEEEtran}
\bibliography{refs}

\end{document}